\documentclass[]{article}
\usepackage[letterpaper]{geometry}
\usepackage{mtsummit2015}
\usepackage{times}
\usepackage{url}
\usepackage{latexsym}
\usepackage{natbib}
\usepackage{layout}
\usepackage{color}
\usepackage[frozencache]{minted}

\usepackage{marginnote}

\usepackage{listings}
\usepackage[dvipsnames]{xcolor}

\newcommand\YAMLcolonstyle{\color{red}\mdseries}
\newcommand\YAMLkeystyle{\color{black}\bfseries}
\newcommand\YAMLvaluestyle{\color{blue}\mdseries}

\makeatletter

\newcommand\lstlanguage@yaml{yaml}

\expandafter\expandafter\expandafter\lstdefinelanguage
\expandafter{\lstlanguage@yaml}
{
  keywords={true,false,null,y,n},
  keywordstyle=\color{darkgray}\bfseries,
  basicstyle=\YAMLkeystyle,                                 
  sensitive=false,
  comment=[l]{\#},
  morecomment=[s]{/*}{*/},
  commentstyle=\color{purple}\ttfamily,
  stringstyle=\YAMLvaluestyle\ttfamily,
  moredelim=[l][\color{orange}]{\&},
  moredelim=[l][\color{magenta}]{*},
  moredelim=**[il][\YAMLcolonstyle{:}\YAMLvaluestyle]{:},   
  morestring=[b]',
  morestring=[b]",
  literate =    {---}{{\ProcessThreeDashes}}3
                {>}{{\textcolor{red}\textgreater}}1     
                {|}{{\textcolor{red}\textbar}}1 
                {\ -\ }{{\mdseries\ -\ }}3,
}

\lst@AddToHook{EveryLine}{\ifx\lst@language\language@yaml\YAMLkeystyle\fi}
\makeatother

\newcommand\ProcessThreeDashes{\llap{\color{cyan}\mdseries-{-}-}}

\begin{document}

\title{\bf XNMT: The eXtensible Neural Machine Translation Toolkit}  

\author{\name{\bf Graham Neubig} \hfill 
        \addr{Carnegie Mellon University, Pittsburgh, USA}
\AND
       \name{\bf Matthias Sperber} \hfill
        \addr{Karlsruhe Institute of Technology, Karlsruhe, Germany}
\AND
       \name{\bf Xinyi Wang, Matthieu Felix, Austin Matthews, Sarguna Padmanabhan, Ye Qi, Devendra Singh Sachan} \hfill
       \addr{Carnegie Mellon University, Pittsburgh, USA}
\AND
       \name{\bf Philip Arthur} \hfill
       \addr{Nara Institute of Science and Technology, Nara, Japan}
\AND
       \name{\bf Pierre Godard} \hfill
       \addr{LIMSI/CNRS/Universit\'e Paris-Saclay, Orsay, France}
\AND
       \name{\bf John Hewitt} \hfill
       \addr{University of Pennsylvania, Philadelphia, USA}
\AND
       \name{\bf Rachid Riad} \hfill
       \addr{ENS/CNRS/EHESS/INRIA, Paris, France}
\AND
       \name{\bf Liming Wang} \hfill
       \addr{University of Illinois, Urbana-Champaign, USA}
}

\maketitle
\pagestyle{empty}

\begin{abstract}
This paper describes XNMT, the eXtensible Neural Machine Translation toolkit.
XNMT distinguishes itself from other open-source NMT toolkits by its focus on modular code design,
with the purpose of enabling fast iteration in research and replicable, reliable results.
In this paper we describe the design of XNMT and its experiment configuration system, and
demonstrate its utility on the tasks of machine translation, speech recognition, and multi-tasked machine translation/parsing.
XNMT is available open-source at \url{https://github.com/neulab/xnmt}.
\end{abstract}

\section{Introduction}
Due to the effectiveness and relative ease of implementation, there is now a proliferation of toolkits for neural machine translation \citep{Kalchbrenner2013,Sutskever2014,Bahdanau2014}, as many as 51 according to the tally by \texttt{nmt-list}.\footnote{\url{https://github.com/jonsafari/nmt-list}}
The common requirements for such toolkits are speed, memory efficiency, and translation accuracy, which are essential for the use of such systems in practical translation settings.
Many open source toolkits do an excellent job at this to the point where they can be used in production systems (e.g. OpenNMT\footnote{\url{http://opennmt.net}} is used by Systran \citep{crego2016systran}).

This paper describes XNMT, the eXtensible Neural Machine Translation toolkit, a toolkit that optimizes not for efficiency, but instead for ease of use in practical research settings.
In other words, instead of only optimizing time for training or inference, XNMT aims to reduce the time it takes for a researcher to turn their idea into a practical experimental setting, test with a large number of parameters, and produce valid and trustable research results.
Of course, this necessitates a certain level of training efficiency and accuracy, but XNMT also takes into account a number of considerations, such as those below:
\begin{itemize}
\setlength{\itemsep}{0.5pt}
\item XNMT places a heavy focus on modular code design, making it easy to swap in and out different parts of the model. Ideally, implementing research prototypes with XNMT involves only few changes to existing code.
\item XNMT is implemented in Python, the de facto standard in the research community.
\item XNMT uses DyNet \citep{Neubig2017} as its deep learning framework. DyNet uses dynamic computation graphs, which makes it possible to write code in a very natural way, and benefit from additional flexibility to implement complex networks with dynamic structure, as are often beneficial in natural language processing. Further benefits include transparent handling of batching operations, or even removing explicit batch handling and relying on autobatching for speed-up instead.
\item XNMT of course contains standard NMT models, but also includes functionality for optimization using reinforcement learning \citep{ranzato2015sequence} or minimum risk training \citep{shen-EtAl:2016:P16-1}, flexible multi-task learning \citep{dai2015semi}, encoders for speech \citep{chan2016listen}, and training and testing of retrieval-based models \citep{huang2013learning}.
\end{itemize}

In the remainder of the paper, we provide some concrete examples of the design principles behind XNMT, and a few examples of how it can be used to implement standard models.

\section{Model Structure and Specification}

\subsection{NMT Design Dimensions: Model, Training, and Inference}
\label{sec:nmtdesign}

When training an NMT system there are a number of high-level design decisions that we need to make: what kind of model do we use? how do we test this model? at test time, how do we generate outputs?
Each of these decisions has a number of sub-components.

For example, when specifying our model, if we are using a standard attentional model such as that defined by Bahdanau et al. \citep{Bahdanau2014}, we must at least decide:
\begin{description}
\setlength{\itemsep}{0.5pt}
\item[Input Data Format:] Do we use plain text or structured data such as trees?
\item[Embedding Method:] Do we lookup words in a table or encode their characters or other units?
\item[Encoder:] Do we use bidirectional LSTMs, convolutional nets, self attention?
\item[Decoder:] Do we use standard LSTM-based word-by-word decoders or add tricks such as memory, syntax, or chunks?
\item[Attention Method:] Do we use multi-layer perceptrons or dot-products, or something more complicated?
\end{description}

\noindent
When specifying the training regimen, there are also choices, including:
\begin{description}
\setlength{\itemsep}{0.5pt}
\item[Loss Function:] Do we use maximum likelihood or a sequence-based training criterion such as REINFORCE or minimum-risk training?
\item[Batching:] How many sentences in a mini-batch, and do we sort by length before batching?
\item[Optimizer:] What optimization method do we use to update our parameters?
\item[Stopping Criterion:] How do we decide when to stop training?
\end{description}

\begin{figure}[t!]
\centering
  \begin{minted}[mathescape,
                 numbersep=5pt,
                 gobble=0,
                 frame=lines,
                 framesep=2mm]{yaml}
mini_experiment: !Experiment # top of experiment hierarchy
  exp_global: !ExpGlobal # global (default) experiment settings
    model_file: examples/output/{EXP}.mod
    log_file: examples/output/{EXP}.log
    default_layer_dim: 512
    dropout: 0.3
  model: !DefaultTranslator # attentional seq2seq model
    src_reader: !PlainTextReader
      vocab: !Vocab {vocab_file: examples/data/train.ja.vocab}
    trg_reader: !PlainTextReader
      vocab: !Vocab {vocab_file: examples/data/train.en.vocab}
    src_embedder: !SimpleWordEmbedder {} # {} indicates defaults
    encoder: !BiLSTMSeqTransducer
      layers: 1
    attender: !MlpAttender {}
    trg_embedder: !SimpleWordEmbedder
      emb_dim: 128 # if not set, default_layer_dim is used
    decoder: !MlpSoftmaxDecoder
      layers: 1
      bridge: !CopyBridge {}
  train: !SimpleTrainingRegimen # training strategy
    run_for_epochs: 20
    batcher: !SrcBatcher
      batch_size: 32
    src_file: examples/data/train.ja
    trg_file: examples/data/train.en
    dev_tasks: # what to evaluate at every epoch
      - !LossEvalTask
        src_file: examples/data/dev.ja
        ref_file: examples/data/dev.en
  evaluate: # what to evaluate at the end of training
  - !AccuracyEvalTask
    src_file: examples/data/test.ja
    ref_file: examples/data/test.en
    eval_metrics: bleu
\end{minted}
\vspace{-6mm}
\caption{Example configuration file \label{yaml_example_1}}
\end{figure}

\noindent
And in inference, there are also options:
\begin{description}
\setlength{\itemsep}{0.5pt}
\item[Search Strategy:] Do we perform greedy search? beam search? random sampling? What are the parameters of each?
\item[Decoding Time Score Adjustment:] At decoding time, do we do something like length normalization to give longer hypotheses higher probability?
\end{description}

Within XNMT, effort is made to encapsulate all of these design decisions in Python classes, making it possible for a researcher who wants to experiment with new alternatives to any one of these decisions to implement a new version of the class and compare it with many other similar alternatives.

\subsection{YAML Model Specification}

In order to specify experimental settings, XNMT uses configuration files in YAML\footnote{http://yaml.readthedocs.io/en/latest/example.html} format, which provides an easy-to-read, Python-like syntax.
An example of such a file, demonstrating how it is possible to specify choices along the various design dimensions in \S\ref{sec:nmtdesign} is shown in Figure~\ref{yaml_example_1}.%
\footnote{For many of these parameters XNMT has reasonable defaults, so the standard configuration file is generally not this verbose.}
As shown in the example, XNMT configuration files specify a hierarchy of objects, with the top level always being an \texttt{Experiment} including specification of the model, training, and evaluation, along with a few global parameters shared across the various steps.

One thing that is immediately noticeable from the file is the \texttt{!} syntax, which allows to directly specify Python class objects inside the YAML file.
For any item in the YAML hierarchy that is specified in this way, all of its children in the hierarchy are expected to be the arguments to its constructor (the Python  method).
So for example, if a user wanted to test create a method for convolutional character-based encoding of words \citep{zhang2015character} and see its result on machine translation, they would have to define a new class \texttt{ConvolutionalWordEmbedder(filter\_width, embedding\_size=512)}, implement it appropriately, then in the YAML file replace the \texttt{src\_embedder:} line with
\begin{figure}[h!]
  \begin{minted}[mathescape,
                 numbersep=5pt,
                 gobble=0,
                 frame=lines,
                 framesep=2mm]{yaml}
src_embedder: !ConvolutionalWordEmbedder
  filter_width: 3
  embedding_size: 512
\end{minted}
\vspace{-6mm}
\end{figure} \\
optionally omitting \texttt{embedding\_size:} if the defaults are acceptable.

As may be evident from the example, this greatly helps extensibility for two reasons: (1) there is no passing along of command line arguments or parsing of complex argument types necessary. Instead, objects are simply configured via their Python interface as given in the code, and newly added features can immediately be controlled from the configuration file without extra argument handling. (2) Changing behavior is as simple as adding a new Python class, implementing the required interface, and requesting the newly implemented class in the configuration file instead of the original one.

\subsection{Experimental Setup and Support}

As Figure \ref{yaml_example_1} demonstrates, XNMT supports the basic functionality for experiments described in \S\ref{sec:nmtdesign}.
In the example, the model specifies the input data structure to be plain text (\texttt{PlainTextReader}), word embedding method to be a standard lookup-table based embedding (\texttt{SimpleWordEmbedder}), encoder to be a bidirectional LSTM (\texttt{BiLSTMSeqTransducer}), attender to be a multi-layer perceptron based attention method (\texttt{MlpAttender}), and the decoder to use a LSTM with a MLP-based softmax (\texttt{MlpSoftmaxDecoder}).
Similarly, in the \texttt{training:} and \texttt{evaluate:} subsections, the training and evaluation parameters are set as well.

In addition to this basic functionality, XNMT provides a number of conveniences to support efficient experimentation:
\begin{description}
\setlength{\itemsep}{0.5pt}
\item[Named experiments and overwriting:] All experiments are given a name, in this case \texttt{mini\_experiment}. The \texttt{\{EXP\}} strings in the configuration file are automatically overwritten by this experiment name, which makes it possible to easily distinguish between log files or model files from different experiments.
\item[Multiple experiments and sharing of parameters:] It is also possible to specify multiple experiments in a single YAML file. To do so, we simply define multiple top-level elements of the YAML file. These multiple experiments can share settings through YAML anchors, where one experiment can inherit the settings from another, only overwriting the relevant settings that needs to be changed.

\item[Saving configurations:] In order to aid reproducibility of experiments, XNMT dumps the whole experiment specification when saving a model. Thus, experiments can be re-run by simply opening the configuration file associated with any model.
\item[Re-starting training:] A common requirement is loading a previously trained model, be it for fine-tuning on different data, tuning decoding parameters, or testing on different data. XNMT allows this by re-loading the dumped configuration file, overwriting a subset of the settings such as file paths, decoding parameters, or training parameters, and re-running the experiment. An example is shown in Figure~\ref{yaml_example_3}, which decodes on a different test set by turning training off, and rewriting the specification of the evaluation set.
\item[Random parameter search:] Often we would like to tune parameter values by trying several different configurations. Currently XNMT makes it possible to do so by defining a set of parameters to evaluate and then searching over them using random search. In the future, we may support other strategies such as Bayesian optimization or enumeration.
\end{description}


\begin{figure}[tb]
\centering
  \begin{minted}[mathescape,
                 numbersep=5pt,
                 gobble=0,
                 frame=lines,
                 framesep=2mm]{yaml}
tied_exp: !Experiment
  ...
  model: !DefaultTranslator
    ..
    trg_embedder: !DenseWordEmbedder
      emb_dim: 128
    decoder: !MlpSoftmaxDecoder
      layers: 1
      bridge: !CopyBridge {}
      vocab_projector: !Ref { path: model.trg_embedder }
\end{minted}
\vspace{-6mm}
\caption{Illustration of referencing mechanism \label{yaml_example_2}}
\end{figure}

\begin{figure}[tb]
\centering
  \begin{minted}[mathescape,
                 numbersep=5pt,
                 gobble=0,
                 frame=lines,
                 framesep=2mm]{yaml}
decode_exp: !Experiment
 load: examples/output/standard.mod
 overwrite:
 - path: exp_global.eval_only
   val: True
 - path: evaluate
   val: !AccuracyEvalTask
     src_file: examples/data/head.ja
     ref_file: examples/data/head.en
     hyp_file: examples/output/{EXP}.test_hyp2
\end{minted}
\vspace{-6mm}
\caption{Illustration of load-and-overwrite mechanism. Here, a trained model is loaded for testing on some particular data. \label{yaml_example_3}}
\end{figure}

\section{Advanced Features}

\subsection{Advanced Modeling Techniques}

XNMT aims to provide a wide library of standard modeling tools of use in performing NMT, or sequence-to-sequence modeling experiments in general.
For example, it has support for speech-oriented encoders \citep{chan2016listen,harwath2016unsupervised} that can be used in speech recognition, preliminary support for self-attentional ``Transformer'' models \citep{vaswani2017attention}.
It also has the ability to perform experiments in retrieval \citep{huang2013learning} instead of sequence generation.

\subsection{Parameter Sharing and Multi-task Learning}

Modern deep learning architectures often include parameter sharing between certain components. For example, tying the output projection matrices and embeddings has been proposed by \citet{press2016using}. While it would be possible to develop a specialized component to achieve this, XNMT features a referencing mechanism that allows simply tying the already existing components (Figure~\ref{yaml_example_2}). References are created by specifying the path of the object to which they point, and result in the exact same object instance being used in both places. The only requirement is for the object's interface to be compatible with both usages, which is usually easily achieved using Python's duck typing coding paradigm. 

This component sharing is also very useful in multi-task training paradigms, where two tasks are trained simultaneously and share some or all of their component parts.
This multi-task training can be achieved by replacing the \texttt{SimpleTrainingRegimen} with other regimens specified for multi-task learning, and defining two or more training tasks that use different input data, models, or training parameters.

\subsection{Training and Inference Methods}

XNMT provides several advanced methods for training and inference.
With regards to training, XNMT notably makes it easy to implement other training criteria such as REINFORCE or minimum risk training by defining a separate class implementing the training strategy.
REINFORCE has been implemented, and more training criteria may be added in the near future.
For inference, it is also possible to specify several search strategies (e.g. beam search), along with several length normalization strategies that helps reduce the penalty on long sentences.

\section{Case Studies}

In this section, we describe three case studies of using XNMT to perform a variety of experiments: a standard machine translation experiment (\S\ref{sec:expmt}), a speech recognition experiment (\S\ref{sec:expasr}), and a multi-task learning experiment where we train a parser along with an MT model (\S\ref{sec:expmultitask}).

\subsection{Machine Translation}
\label{sec:expmt}
We trained a machine translation model on the WMT English-German benchmark, using the preprocessed data by Stanford.\footnote{\url{https://nlp.stanford.edu/projects/nmt/}} Our model was a basic 1-layer model with bidirectional LSTM encoder and 256 units per direction, LSTM decoder output projections and MLP attention mechanism all with 512 hidden units. We applied joint BPE of size 32k \citep{sennrich2016bpe}. We also applied input feeding, as well as variational dropout of rate 0.3 to encoder and decoder LSTMs. Decoding was performed with a beam of size 1. Overall, results were similar, with our model achieving a BLEU of 18.26 and \citet{luong2015effective} achieving a BLEU of 18.1. Note that the model by \citet{luong2015effective} is simpler because it does not use BPE and only a unidirectional encoder.


\subsection{Speech Recognition}
\label{sec:expasr}
We performed experiments in a speech recognition task with a simple listen-attend-spell model \citep{chan2016listen}. This model features a 4-layer pyramidal LSTM encoder, subsampling the input sequence by factor 2 at every layer except the first, resulting in an overall subsampling factor of 8. The layer size is set to 512, the target embedding size is 64, and the attention uses an MLP of size 128. Input to the model are Mel-filterbank features with 40 coefficients. For regularization, we apply variational dropout of rate 0.3 in all LSTMs, and word dropout of rate 0.1 on the target side \citep{gal2016theoretically}. For training, we use Adam \citep{kingma2014adam} with initial learning rate of 0.0003, which is decayed by factor 0.5 if no improved in WER is observed. To further facilitate training, label smoothing \citep{szegedy2016rethinking} is applied. For the search, we use beam size 20 and length normalization with the exponent set to 1.5. We test this model on both the Wall Street Journal (WSJ; \citet{paul1992design}) corpus which contains read speech, and the TEDLIUM corpus \citep{rousseau2014enhancing} which contains recorded TED talks. Numbers are shown in Table~\ref{tab:asr_results}. Comparison to results from the literature shows that our results are competitive.
\begin{table}[tb]
\centering
\label{tab:asr_results}
\begin{tabular}{lllll}
Model                           & WSJ dev93 & WSJ eval92 & TEDLIUM dev & TEDLIUM test \\
\hline
XNMT                            & 16.65     & 13.50       & 15.83 & 16.16       \\
\citet{zhang2017very}           & ---       & 14.76      & ---   &       \\
\citet{rousseau2014enhancing}   & ---       & ---        & 15.7  & 17.8
\end{tabular}
\caption{Speech recognition results (WER in \%) compared to a similar pyramidal LSTM model \citep{zhang2017very} and a highly engineered hybrid HMM system \citep{rousseau2014enhancing}.}
\end{table}

\subsection{Multi-task MT + Parsing}
\label{sec:expmultitask}
We performed a multi-task training of a sequence-to-sequence model for parsing  and a machine translation task. The main task is the parsing task, and we followed the general setup in \citep{VinyalsKKPSH15}, but we only used the standard WSJ training data. It is jointly trained with English-German translation system. Since the source side is English sentence for both tasks, they can share the source embedder and source encoder. We also trained a single sequence-to-sequence model for parsing with the same hyperparameters as the multi-task model. As a result, a model trained only on WSJ achieved a test F-score of 81\%, while the multi-task trained model achieved an F-score of 83\%. This experiment was done with very few modifications to existing XNMT multi-task architecture, demonstrating that it is relatively easy to apply multi-tasking to new tasks.

\section{Conclusion}

This paper has introduced XNMT, an NMT toolkit with extensibility in mind, and has described the various design decisions that went into making this goal possible.

\section*{Acknowledgments}

Part of the development of XNMT was performed at the Jelinek Summer Workshop in Speech and Language Technology (JSALT) ``Speaking Rosetta Stone'' project \citep{scharenborg18icassp}, and we are grateful to the JSALT organizers for the financial/logistical support, and also participants of the workshop for their feedback on XNMT as a tool.

Parts of this work were sponsored by Defense Advanced Research Projects Agency Information Innovation Office  (I2O).  Program:    Low  Resource  Languages for   Emergent   Incidents   (LORELEI).   Issued   by DARPA/I2O  under  Contract  No. HR0011-15-C-0114.  The views and conclusions contained in this document  are  those  of  the  authors  and  should  not be  interpreted  as  representing  the  official  policies, either  expressed  or  implied,  of  the  U.S.  Government.   The  U.S.  Government  is  authorized  to  reproduce and distribute reprints for Government purposes notwithstanding  any copyright  notation here on.

\small

\bibliographystyle{apalike}
\bibliography{library}

\end{document}